\documentclass[runningheads]{llncs}
\usepackage{graphicx}
\usepackage{caption}
\usepackage{tikz}
\usepackage{comment}
\usepackage{amsmath,amssymb} % define this before the line numbering.
\usepackage{color}
\usepackage{url}

\usepackage[accsupp]{axessibility}  % Improves PDF readability for those with disabilities.

\usepackage{breakcites}
\usepackage[width=122mm,left=12mm,paperwidth=146mm,height=193mm,top=12mm,paperheight=217mm]{geometry}

\begin{document}
% \renewcommand\thelinenumber{\color[rgb]{0.2,0.5,0.8}\normalfont\sffamily\scriptsize\arabic{linenumber}\color[rgb]{0,0,0}}
% \renewcommand\makeLineNumber {\hss\thelinenumber\ \hspace{6mm} \rlap{\hskip\textwidth\ \hspace{6.5mm}\thelinenumber}}
% \linenumbers
\pagestyle{headings}
\mainmatter
\def\ECCVSubNumber{1131}  % Insert your submission number here

\title{Shape-guided Object Inpainting} % Replace with your title

% INITIAL SUBMISSION 
\begin{comment}
\titlerunning{ECCV-22 submission ID \ECCVSubNumber} 
\authorrunning{ECCV-22 submission ID \ECCVSubNumber} 
\author{Anonymous ECCV submission}
\institute{Paper ID \ECCVSubNumber}
\end{comment}
%******************

% CAMERA READY SUBMISSION
% \begin{comment}
\titlerunning{Shape-guided Object Inpainting}
% If the paper title is too long for the running head, you can set
% an abbreviated paper title here
%
\author{Yu Zeng\inst{1} \and
Zhe Lin\inst{2} \and
Vishal M. Patel\inst{1}}
\authorrunning{Y. Zeng et al.}
% First names are abbreviated in the running head.
% If there are more than two authors, 'et al.' is used.
%
\institute{Johns Hopkins University,  \\
\email{\{yzeng22,vpatel36\}@jhu.edu}
\and
Adobe Research\\
\email{zlin@adobe.com}}
% \end{comment}
%******************
\maketitle

\begin{center}
    \centering
    \includegraphics[width=\textwidth]{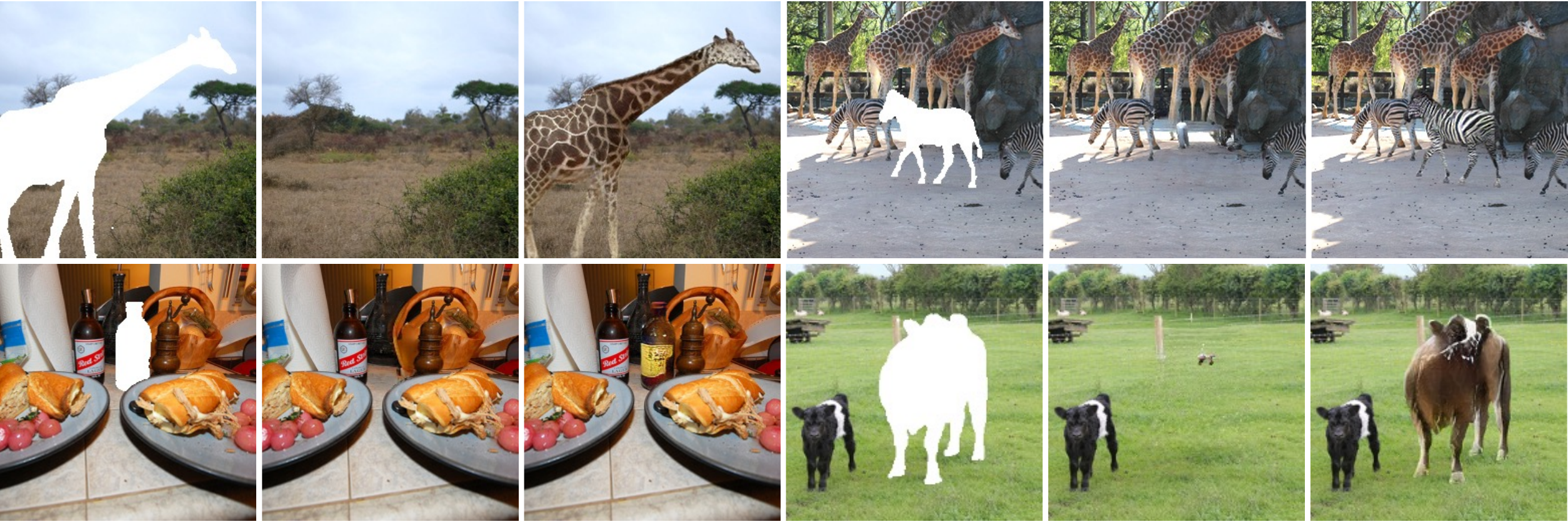}\\
    \scriptsize{\hfill{Input} \hfill\hfill  {CoModGAN~\cite{zhao2021comodgan}} \hfill\hfill  {Ours} \hfill\hfill {Input} \hfill\hfill {CoModGAN~\cite{zhao2021comodgan}} \hfill\hfill {Ours} \hfill}
    \captionof{figure}{We explore a new image inpainting task: shape-guided object inpainting. Given an incomplete input image, existing inpainting methods fill the hole by generating background, while our method generates an object. }
    \label{teaser}
\end{center}%

\begin{abstract}
Previous works on image inpainting mainly focus on inpainting background or partially missing objects, while the problem of inpainting an entire missing object remains unexplored. 
This work studies a new image inpainting task,~\ie shape-guided object inpainting. Given an incomplete input image, the goal is to fill in the hole by generating an object based on the context and implicit guidance given by the hole shape. 
Since previous methods for image inpainting are mainly designed for background inpainting, they are not suitable for this task. 
Therefore, we propose a new data preparation method and a novel Contextual Object Generator (CogNet) for the object inpainting task. 
On the data side, we incorporate object priors into training data by using object instances as holes. The CogNet has a two-stream architecture that combines the standard bottom-up image completion process with a top-down object generation process. A predictive class embedding module bridges the two streams by predicting the class of the missing object from the bottom-up features, from which a semantic object map is derived as the input of the top-down stream. 
Experiments demonstrate that the proposed method can generate realistic objects that fit the context in terms of both visual appearance and semantic meanings. Code can be found at the project page \url{https://zengxianyu.github.io/objpaint}

\end{abstract}

\section{Introduction}
Image inpainting (a.k.a. image completion), which aims
to fill missing regions of an image, has been an active research topic of computer vision for decades. 
Despite the great progress made in recent years~\cite{lahiri2020prior,suin2021distillation,zhou2021transfill,yi2020contextual,nazeri2019edgeconnect,iizuka2017globally,liu2018image,xiong2019foreground,ren2019structureflow,liao2021image,xiao2019cisi,yu2020region,yang2020learning,yang2017high,wangimage,pathak2016context,song2018contextual,ren2019structureflow}, image inpainting remains a challenging problem due to its inherent ambiguity and the complexity of natural images. Therefore, various guided inpainting methods have been proposed that exploit external guidance information such as examplar~\cite{kwatra2005texture,zhao2019guided,zhou2021transfill}, sketches~\cite{liu2021deflocnet,yang2020deep,jo2019sc,portenier2018faceshop,yu2019free}, label maps~\cite{ardino2021semantic},~\etc. However, previous work on image inpainting mainly focuses on inpainting background or partially missing objects. The problem of inpainting an entire missing object is still unexplored. In this paper, we study a new guided inpainting task,~\ie shape-guided object inpainting, where the guidance is implicitly given by the object shape. As shown in Fig.~\ref{teaser}, given an incomplete input image, the goal is to generate a new object to fill the hole. It can be used in various practical applications such as object re-generation, object insertion, and object/person anonymization. 
% thanks to the rapid development in image processing techniques and hardware. 
% There are also guided inpainting methods use guidance to improve inpainting results. However these methods are mainly for backgroudn inpainting and are not designed for object generation. In this paper, we study a new guided inpainting task,~\ie shape-guided object inpainting. Given an incomplete input image, the goal is to generate a new object to fill the hole, as shown in Fig.~\ref{teaser}. 

This task has a similar input and output setup to the traditional image inpainting task; both take an incomplete/masked image and the hole mask as input to produce a complete image as output. However, previous methods are mainly designed for background inpainting and are not suitable for this object inpainting task. 
Early patch-based synthesis methods borrow content from the remaining image to fill the hole. These methods are hardly seemed fit for this task as they cannot generate novel content. 
Recent deep generative inpainting methods should be able to inpaint both background and objects, but in practice, they still have a strong bias towards background generation~\cite{katircioglu2020self}. 
% Ideally, the deep generative inpainting methods should be able to inpaint a missing region with both background and objects as they are both reasonable solutions. However, in practice, existing generative inpainting models also are shown to have a strong bias towards background generation~\cite{katircioglu2020self}. 
The reason lies in both the training strategy and the model architecture of previous deep learning based approaches. 
First, previous methods synthesize the training data by simply masking images at random positions with different regions masked at equal probability. Since the appearance of background patches are usually similar to surrounding, it is easier to learn to extend the surrounding background to fill a hole than to generate objects. 
Second, previous methods formulate image inpainting as a bottom-up context-based process that uses stacked convolution layers to propagate context information from the known region to the missing regions. However, object generation is essentially a top-down process: it starts from a high-level concept of the object and gradually hallucinate the concrete appearance centering around the concept. Without any top-down guidance, it is hard to generate a reasonable object of consistent semantic meaning. 

Therefore, in order to find a better solution, we design a new data preparation method and a new generative network architecture for the object inpainting task. On the data side, to overcome the bias towards the background, we incorporate object prior by using object instances as holes in training. For the network architecture, we consider three important goals of object inpainting: 
(1) visual coherency between the appearance of generated and existing pixels; 
(2) semantic consistency within the inpainted region,~\ie the generated pixels should constitute a reasonable object; 
(3) high-level coherency between the generated objects and the context. 
To achieve these goals, we propose a contextual object generator (CogNet) with two-stream network architecture. 
It consists of a bottom-up and top-down stream that models a bottom-up and top-down generation process, respectively. 
The bottom-up stream resembles a typical framework used by previous approaches to achieve appearance coherency. It takes the incomplete image as input and fills the missing region based on contextual information extracted from the existing pixels. 
The bridge between the bottom-up stream is a predictive class embedding (PCE) module. It predicts the class of the missing object based on features from the bottom-up stream to encourage high-level coherency. 
The top-down stream is designed inspired by semantic image synthesis~\cite{isola2017image,park2019semantic} and has a similar framework to it. 
It aims to hallucinate class-related object features based on a semantic object map obtained by combining the predicted class and the hole mask. Since the features at all object pixels are generated from the same class label, their semantic consistency can be ensured.

In summary, our contributions are as follows:
\begin{itemize}
\item We explore a new guided image inpainting task,~\ie shape-guided object inpainting. 
\item We propose a new data preparation method and a novel Contextual Object Generator (CogNet) model for object inpainting. 
\item Experiments demonstrate that the proposed method is effective for the task and achieves superior performance against state-of-the-art inpainting models finetuned for the task. 
\end{itemize}

\section{Related Work}
\subsection{Image Inpainting}
Conventional image inpainting methods fill the holes by borrowing existing content from the known region. Patch-based methods search well-matched patches from the known part in the input image as replacement patches to fill in the missing region. Efros~\etal~\cite{efros1999texture} propose a non-parametric sampling method for texture synthesis method that can synthesize images by sampling patches from a texture example. It can be applied for hole-filling through constrained texture synthesis. Drori~\etal~\cite{drori2003fragment} propose to iteratively fill missing regions from high to low confidence with similar patches. Barnes~\etal~\cite{barnes2009patchmatch} propose a randomized algorithm for quickly finding matched patches for filling missing regions in an image. Diffusion-based methods propagate local image appearance surrounding the missing region based on the isophote direction field. Bertalmio~\etal~\cite{10.1145/344779.344972} propose to smoothly propagate information from the surrounding areas in the isophotes direction to fill the missing regions. Ballester~\etal~\cite{ballester2001filling} propose to jointly interpolate the image gray-levels and gradient/isophotes directions to smoothly extend the isophote lines into the holes. 
These methods cannot generate entirely new content that does not exist in the input image. 

In recent years, driven by the success of deep generative models, extensive research efforts have been put into data-driven deep learning based approaches. This branch of work usually formulates image completion as an image generation problem conditioned on the existing pixels in known regions. 
They can generate plausible new content and have shown significant improvements in filling holes in complex images. 
The first batch of deep learning based approaches only works on square holes. Iizuka~\etal~\cite{iizuka2017globally} propose to use two discriminators to train a conditional GAN to make the inpainted content both locally and globally consistent. Yu~\etal~\cite{yu2018generative} propose contextual attention to explicitly utilize surrounding image features as references in the latent feature space. Zeng~\etal~\cite{zeng2019learning} propose to use region affinity from high-level features to guide the completion of missing regions in low-level features. Later on, the research effort has shifted to image completion with irregular holes. Liu~\etal~\cite{liu2018image} use collect estimated occlusion/dis-occlusion masks between two consecutive frames of videos and use them to generate holes and propose partial convolution to exploit information from the known region more efficiently. Yu~\cite{yu2019free} generate free-form masks by simulating random strokes. They generalize partial convolution to gated convolution that learns to select features for each channel at each spatial location across all layers. Zeng~\etal~\cite{zeng2020high} use object-shaped holes to simulate real object removal cases and propose an iterative inpainting method with a confidence feedback mechanism. 
The above deep learning based methods mainly focus on background inpainting. In training, images are masked at random positions, resulting in a bias towards background as background is usually more predictable in most images. In addition, some methods use attention mechanisms to explicitly borrow patches/features from known regions~\cite{yu2018generative,yu2019free,zeng2019learning,zeng2020high,zhang2019residual,liu2019coherent} as in the conventional methods, which can be seen as background prior and will further encourage the tendency to generate background. Some previous works on deep learning based inpainting have touched on topics related to object inpainting. Xiong~\etal~\cite{xiong2019foreground} propose a foreground-aware image inpainting system by predicting the contour of salient objects. Ke~\etal~\cite{ke2021occlusion} propose an occlusion aware inpainting method to inpaint partially missing objects in videos. These methods mainly focus on inpainting partially missing objects. 

\subsection{Guided Image Inpainting}
Some works attempt to allow users to provide more guidance to reduce the ambiguity of image inpainting and improve the results. Many types of guidance have been explored, such as examplar images, sketches, label maps, text. 
Yu~\etal~\cite{yu2019free} propose DeepFillV2, which can perform sketch-guided image inpainting of general images as well as face images. 
Park~\cite{jo2019sc} explore face inpainting with sketch and color strokes as guidance. 
Zhang~\etal~\cite{zhang2020text} propose to inpaint the missing part of an image according to text guidance provided by users. Ardino~\etal~\cite{ardino2021semantic} propose to use label maps as guidance for image inpainting. Although the guided inpainting methods \cite{zhang2020text} and \cite{ardino2021semantic} might be able to generate an entire new object if the text or label map about the object are given as guidance, they require the users to provide the external guidance explicitly. In comparison, our method only takes the incomplete image and hole mask as input. 

% \subsection{Conditional Image Generation}
% \textcolor{red}{Earlier image inpainting methods rely on existing content to fill the holes. Diffusion-based methods [8,10] propagate neighboring appearances to the target holes, but they often generate significant artifacts when the holes are large or texture variation is severe. Patch-based methods [17,25,9] search for most similar patches from valid regions to complete missing regions. Drori et al. [16] propose to iteratively fill missing regions from high to low confidence with similar patches. Although they also use a map to determine the region to fill in each iteration, the map is predefined based on spatial distances from unknown pixels to their closest valid pixels. The above methods use real image patches sampled from the input to fill the holes and can often generate high-quality results. However, they lack high-level structural understanding and cannot generate entirely new content that does not exist in the input image. Thus, their results may not be semantically consistent to regions surrounding the holes.}

\subsection{Semantic Image Synthesis}
Semantic image synthesis is a sub-class of conditional image generation which aims to generate photo-realistic images from user-specified semantic layouts. It was first introduced by Isola~\etal~\cite{isola2017image}, who proposed an image-to-image translation framework, called Pix2Pix, to generate images from label maps or edge maps. 
Zhu~\etal~\cite{zhu2017unpaired} propose CycleGAN to allow training an image translation model on unpaired data with a cycle consistency constraint. Park~\etal~\cite{park2019semantic} propose spatially-adaptive normalization for semantic image synthesis, which modulates the activations using semantic layouts to propagate semantic information throughout the network. Chen~\etal~\cite{chen2017photographic} propose cascaded refinement networks and use perceptual losses for semantic image synthesis. Wang~\etal~\cite{wang2018high} propose Pix2PixHD which improves the quality of synthesized images using feature matching losses, multiscale discriminators and an improved generator. Our method takes inspiration from semantic image synthesis methods to design the top-down stream of the contextual object generator. Unlike semantic image synthesis, where the semantic layouts or label maps are known, our semantic object maps are derived by combining the predicted class and the hole mask.

\subsection{Background-based Object Recognition}
Object recognition is a task to categorize an image according to the visual
contents. In recent years, the availability of large-scale datasets and powerful computers made it possible to train deep CNNs, which achieved a breakthrough success for object recognition~\cite{krizhevsky2012imagenet}. 
Normally, an object recognition model categorizes an object primarily by recognizing the visual patterns in the foreground region. However, recent research has shown that a deep network can produce reasonable object results with only background available. Zhu~\etal~\cite{zhu2016object} find that the AlexNet model~\cite{krizhevsky2012imagenet} trained on pure background without objects achieves highly reasonable recognition performance that beats human recognition in the same situations. 
Xiao~\etal~\cite{xiao2020noise} analyze the performance of state-of-the-art architectures on object recognition with foreground removed in different ways. It is reported that the models can achieve over 70\% test accuracy in a no-foreground setting where the foreground objects are masked. These works aim to predict only the class of an object from background. In this paper, we show that the entire object can be generated based on the background. 

\section{Method}
Given an input image with missing regions, our goal is to fill the missing region with generated objects. We take a data-driven approach based on generative adversarial networks (GANs)~\cite{goodfellow2014generative,radford2015unsupervised,karras2017progressive,brock2018large,karras2019style,karras2020analyzing}. A  contextual object generator is designed to generate objects based on the context that not only fit the known region and of reasonable semantic meanings. 
% We design a contextual object generator $g(\cdot)$ that takes an incomplete image $x'$ and the mask $m$ of missing region as input, and produces the completed image $\hat{x}=g(x')\odot m + x'$. 
The generator is jointly trained with a discriminator on a synthetic dataset obtained by masking object regions in real images. We use the discriminator proposed in \cite{karras2019style,karras2020analyzing}. In what follows, we introduce our data acquisition approach in Sec.~\ref{sec:data} and network architecture of the generator in Sec.~\ref{sec:gen}. 

\subsection{Data Preparation}
\label{sec:data}
\begin{figure}
\begin{center}
    \centering
    \includegraphics[width=\textwidth]{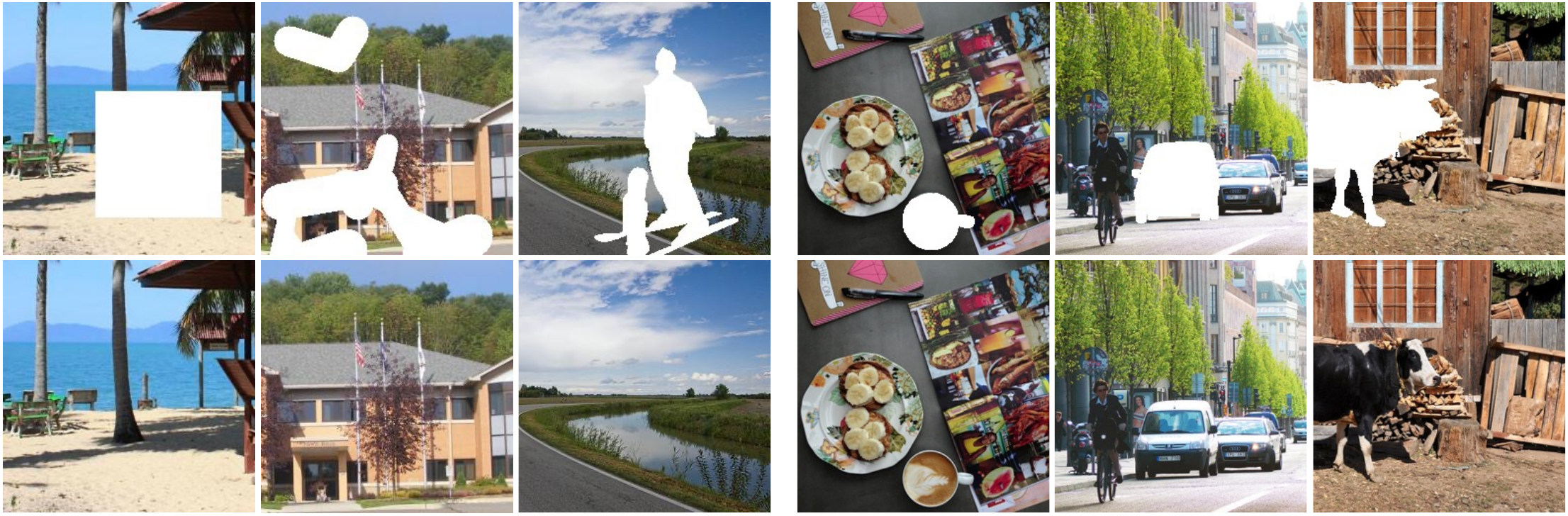}\\
% \scriptsize{\hfill{Images} \hfill\hfill  {Sketches} \hfill\hfill  {Results} \hfill\hfill {Images} \hfill\hfill {Sketches} \hfill\hfill {Results} \hfill}
\scriptsize{\hfill\hfill  {Previous approaches} \hfill\hfill  {\textcolor{white}{ious app}Ours\textcolor{white}{roaches}} \hfill\hfill}
\caption{Top: input. Bottom: original image and ground-truth. Previous deep learning based inpainting methods generate training data by masking at random positions, which results in a bias towards background generation. We propose to incorporate object prior into training data by masking object instances. }
    \label{fig0}
    \vspace{-10pt}
\end{center}%
\end{figure}
Most deep learning based image inpainting methods prepare data by masking images at random positions using synthetic masks obtained by drawing random rectangles~\cite{zeng2019learning,yu2018generative,yang2017high}, brush strokes or from a fixed set of irregular masks~\cite{liu2018image,zeng2020high,liu2020rethinking}. Paired training data $\{(x',m),x\}$ can be formed by taking the masked image $x'=x\odot m$ and mask $m$ as input with the original image $x$ as ground-truth. 
This data synthesis pipeline can generate a very large dataset for training a powerful deep model capable of completing large holes and dealing with complex scenes. 
Although this random masking process produces diverse data with masks on both background and object regions, the trained model often has a strong tendency to generate background as background is more common and easier to predict than objects~\cite{joung2012reliable,katircioglu2019self}. 
In this work, since we aim to train an image completion model to generate objects, the random masking process is not suitable. Therefore, we design a new data synthesis method that incorporates the object prior into training data by using object instances as holes. 
For an image $x$, its instance segmentation $\{m^i, y^i\}_{i=1}^c$ can be obtained by manual annotation or using segmentation models, where $m^i, y^i$ are the mask and class of each object instance, $c$ denotes the number of instances. Then $c$ training samples $\{(x'^i,m^i),x\}_{i=1}^c$ can be constructed by masking the image $x$ with each instance mask: $x'^i=x\odot m^i$. 
There exist datasets such as COCO~\cite{lin2014microsoft} with manually annotated segmentation masks, which can be used to construct high-quality training samples for object-based image completion. However, these datasets are limited in size and are not sufficient for representing the complexity of objects in natural images. To obtain larger and more diverse training data, we can use instance segmentation models to automatically label a larger dataset with instance masks complementary to the manually annotated segmentation datasets. Although the automatically annotated masks are less accurate, they still cover most object regions and thus can provide a reasonable object prior. 
Fig.~\ref{fig0} compares our training samples with object instances as holes and the randomly generated training samples used in previous approaches. 

\subsection{Network Architecture}
\label{sec:gen}
\begin{figure}
\begin{center}
    \centering
    \includegraphics[width=\textwidth]{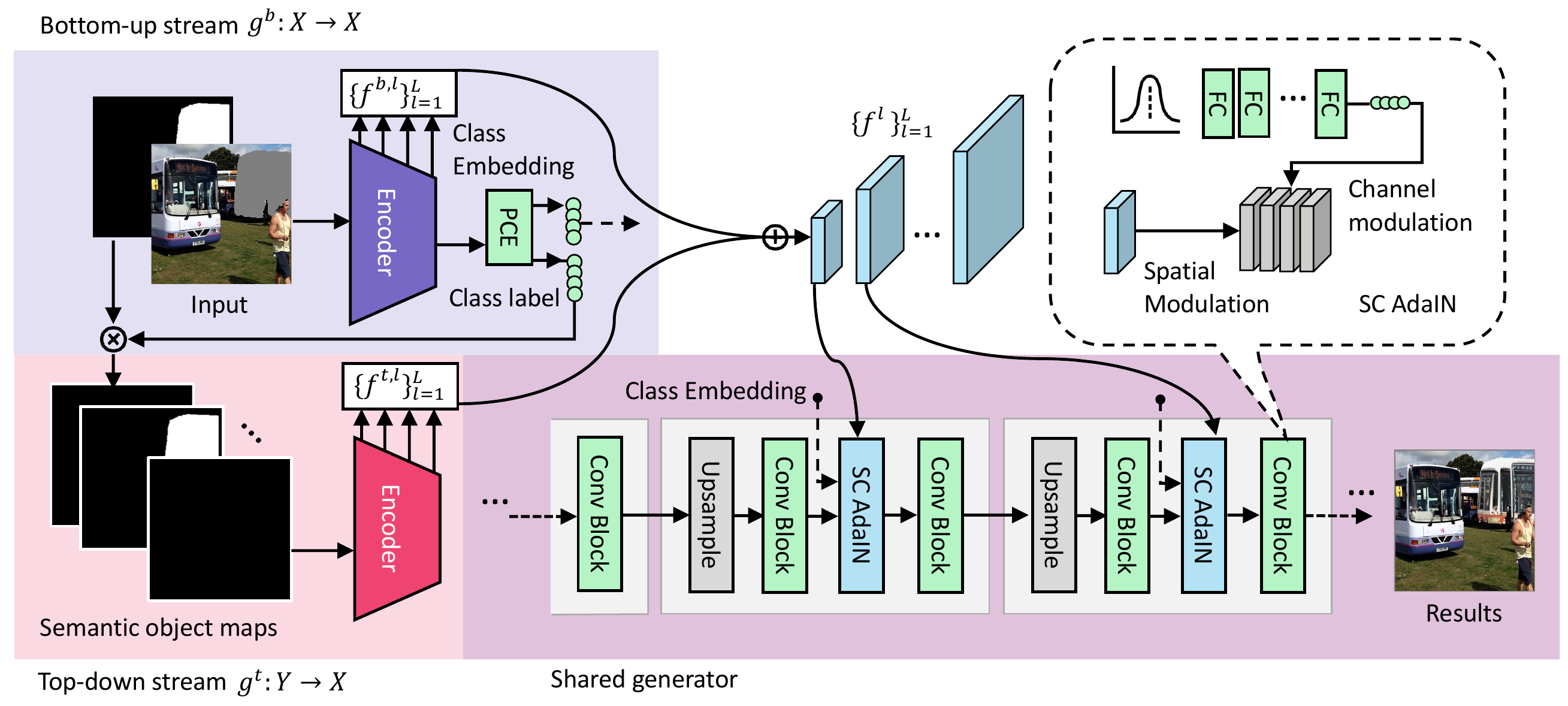}
    % \scriptsize{\hfill{Images} \hfill\hfill  {Sketches} \hfill\hfill  {Results} \hfill\hfill {Images} \hfill\hfill {Sketches} \hfill\hfill {Results} \hfill}
\caption{Illustration of the two-stream network architecture. It consists of a bottom-up stream and a top-down stream. The bottom-up stream models the standard image inpainting process, which takes an incomplete image as input to produce a complete image. The predictive class embedding (PCE) predicts the object class label based on features from the bottom-up stream and embeds it into an embedding vector. The top-down stream generates an image conditioned on the semantic object map. The two streams share the same generator. }
    \label{fig1}
    \vspace{-10pt}
\end{center}%
\end{figure}
In this section, we present the network architecture of the proposed contextual object generator (CogNet). 
Unlike the traditional image completion task, which only focuses on the consistency of the inpainted region and the context, object-based image completion also requires the inpainted content to be an object of semantic meanings. 
Previous network architectures for image completion are mainly designed as a bottom-up process to propagate information from known regions to missing regions. The generated content can blend naturally into the context but rarely resemble an object due to the lack of top-down guidance. 
% This object generation problem comes with two constraints: (1) similar to traditional image completion cases, the feature(?) (~\eg color, illumination,~\etc) of the generated content should be consistent with the surrounding area; (2) the generated content should be an object of semantic meanings. Previous network architectures proposed for the traditional image completion cases are mainly designed as a bottom-up process to propagate information from the known region to missing region. The generated content can blend naturally into the context, however, rarely resemble an object due to the lack of top-down guidance. 
To solve this problem, we design a two-stream architecture that combines the traditional image inpainting framework with a top-down object generation process inspired by the semantic image synthesis task~\cite{isola2017image,park2019semantic}. The overall structure is shown in Fig.~\ref{fig1}. Each stream has an independent encoder that takes input from the corresponding domains and interacts with each other through the shared generator. 

\subsubsection{Bottom-up Process}
The bottom-up stream $g^b$ follows the standard design of an image inpainting model. It takes an incomplete RGB image $x' \in X$ and the hole mask $m$ as input and produce an inpainted RGB image $\hat{x} \in X$,~\ie $g^b:X\rightarrow X$. 
Given an incomplete input image, the encoder extracts hierarchical features from the raw pixels of the known region. It consists of a sequence of $L$ convolutional blocks with a $2\times$ downsample operator between every two consecutive blocks. For an input of size $N\times N$, the encoder produces a series of feature maps $\{ f^{b,l} \}_{l=0}^{L-1}$ of various scales, where each feature map $f^{b,l}$ is of size $\frac{N}{2^l}$. Then the multi-scale feature maps $\{ f^{b,l} \}$ are used to modulate the generator features of the corresponding scale through the spatial-channel adaptive instance normalization (SC AdaIN) layers. 

\subsubsection{Predictive Class Embedding}
The bottom-up stream can capture the environmental factor that affects the object's appearance, such as color, illumination, and style. However, the class-related information is still missing. 
As recent studies~\cite{xiao2020noise,zhu2016object} have indicated, models can achieve reasonable object recognition performance by relying on the background alone. Based on this observation, we propose a predictive class embedding module to map the background features into object class embeddings by learning a background-based object recognition model. 
First, the feature $f^{b,L-1}$ of the last block of the encoder is reshaped and transformed by a fully connected layer into a feature vector $h$. Then a linear classifier is trained to predict the object class given $h$ as input by minimizing $\mathcal{L}_c$:
\begin{equation}
\label{eq:loss_cls}
\mathcal{L}_c = \sum_i -t_i \log \hat{t}_i, \mbox{ where } \hat{t} = \sigma (W^c h)_i,
\end{equation}
where $t$ is the one-hot encoding of the true class label; $W^c$ is the weight of the linear classifier; $\sigma$ represents the softmax function; $\hat{t}$ represents the predicted class label. $h$ can be seen as an embedding of the predicted class and is also passed into the SC AdaIN layers. 

\subsubsection{Top-down Process}
In most images, the appearance of the objects is less predictable from the context than background. Hence the bottom-up process is less effective for object-based image completion. 
Therefore, we design a top-down stream to allow the model to hallucinate appearance features from semantic concepts for object generation. 
The top-down stream $g^t:Y\rightarrow X$ is designed inspired by semantic image synthesis methods,~\ie generating image content from semantic layout. 
Different from standard semantic image synthesis where the label maps are known, the top-down stream generated an RGB image based on the semantic object maps derived from the predicted class. 
More specifically, given the predicted class $\hat{t}$, a semantic object map $y \in Y$ can be derived by combining the predicted class and the hole mask $m$:
\begin{equation}
y_i = \hat{t}_i \cdot m,
\end{equation}
where $y_i$ represents the semantic object map corresponding to the $i$-th class. Then an $L$-layer encoder with a similar structure to the one in the bottom-up stream encodes the semantic object maps into multi-scale feature maps $\{f^{t,l} \}_{l=1}^{L}$. These feature maps will be used to modulate the generator feature maps through SC AdaIN layers to provide spatial aware class-related information to the generator. 

\subsubsection{SC AdaIN}
\begin{figure}
\begin{center}
    \centering
    \includegraphics[width=\textwidth]{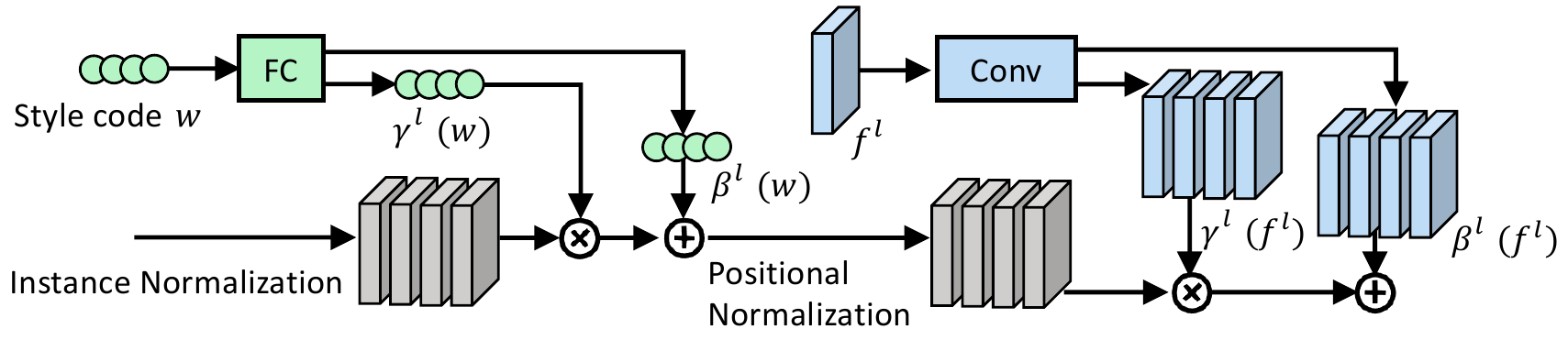}
    % \scriptsize{\hfill{Images} \hfill\hfill  {Sketches} \hfill\hfill  {Results} \hfill\hfill {Images} \hfill\hfill {Sketches} \hfill\hfill {Results} \hfill}
\caption{Illustration of the spatial-channel adaptive instance normalization module. It consists of two steps of normalization and modulation in the channel and spatial dimensions, respectively. }
    \label{fig2}
    \vspace{-10pt}
\end{center}%
\end{figure}
Given the environmental features and class inferred from the background, there still can be many possible object appearances. 
To model the uncertainty in object generation while preserving the information propagated from the encoders, we design the spatial-channel adaptive instance normalization module (SC AdaIN). Fig.~\ref{fig2} illustrates the structure of a SC AdaIN module. 
Given an input image, we obtain the multi-scale feature maps $\{f^{b,l}\}, \{f^{t,l}\}$ from the encoders and sample a random latent code $z \sim \mathcal{N}(0,1)$. Then the latent code is transformed by a fully connected network as in~\cite{karras2020analyzing,karras2019style} and concatenated with the class embedding $h$ into a style code $w$. 
For each scale $l$, we normalize and modulate the generator feature map channel-wise using the encoder features and position-wise using the style code $w$. 
Let $X^l$ denote the generator feature map at scale $l$, the modulated feature map $\hat{X}^l$ is produced as follow,
\begin{equation}
\bar{X}^l_{c,x,y} = \frac{X^l_{c,x,y}-\mu^l_{c}}{\sigma^l_c}\cdot \gamma^l(w)_c + \beta^l(w)_c
\end{equation}
\begin{equation}
\hat{X}^l_{c,x,y} = \frac{\bar{X}^l_{c,x,y}-\bar{\mu}^l_{x,y}}{\bar{\sigma}^l_{x,y}}\cdot \bar{\gamma}^l(f^{b,l}+f^{t,l})_{c,x,y} + \bar{\beta}^l(f^{b,l}+f^{t,l})_{c,x,y}
\end{equation} 
where $\mu^l_c, \sigma^l_c$ are the mean and standard deviation of $X^l$ in channel $c$; $\bar{\mu}_{x,y}, \bar{\sigma}_{x,y}$ are the mean and standard deviation of $\bar{X}^l$ at position $x,y$; $\gamma^l(w), \beta^l(w)$ and $\bar{\gamma}^l(f^l), \bar{\beta}^l(f^l)$ transform the style code $w$ and the encoder feature maps $f^l$ into the modulation parameters at scale $l$.  

\section{Experiment}
\subsection{Implementation Details}
We implement our method and train the model using Python and Pytorch~\cite{NEURIPS2019_9015}. We use the perceptual loss~\cite{johnson2016perceptual}, GAN loss~\cite{gulrajani2017improved}, and the loss in Eqn.~\ref{eq:loss_cls} to train the contextual object generator. The detailed network architectures can be found in the supplementary material. The code will be made publicly available after the paper is published. 
The model is trained on two A100 GPUs. It takes about a week for training. The inference speed at $256\times256$ resolution is 0.05 seconds per image. 
We compare with two state-of-the-art image inpainting methods DeepfillV2~\cite{yu2019free} and CoModGAN~\cite{zhao2021comodgan} and RFR~\cite{li2020recurrent}. Since the original models of the compared methods are trained using random masks, it is not suitable to directly apply the pretrained models for object inpainting. Therefore, to compare with these methods, we train the model on the corresponding dataset using the mask synthesis method described in Sec.~\ref{sec:data}. We evaluate the performance using the metrics FID~\cite{heusel2017gans} and LPIPS~\cite{zhang2018perceptual} as they are the most commonly used metric for assessing the quality of generative models~\cite{lucic2018gans} and image-conditional GANs~\cite{albahar2019guided,huang2018multimodal,shen2019towards}. 

\subsection{Datasets}
We train and evaluate our model on three datasets COCO~\cite{lin2014microsoft}, Cityscapes~\cite{Cordts2016Cityscapes} and Places2~\cite{zhou2017places}, which are commonly used in image inpainting, semantic segmentation, and semantic image synthesis. Note that the segmentation maps are only required in training. In the inference stage, only an input image and hole mask are needed. 
We use the official training split to train and evaluate the models on the official validation split. All images are cropped into $256\times 256$ patches during training and evaluation.  
Cityscape dataset contains segmentation ground truths for objects in city scenes such as roads, lanes, vehicles, and objects on roads. This dataset contains 30 classes collected over different environmental and weather conditions in 50 cities. It provides dense pixel-level annotations for 5,000 images pre-split into training (2,975), validation (500) and test (1,525). Since Cityscapes provides accurate segmentation ground-truth, it can be directly used for training our model. 
COCO dataset is a large-scale dataset designed to represent a vast collection of common objects. This dataset is split into a training split of 82,783 images, a validation split of 40,504 images, and a test split of 40,775 images. 
There are 883,331 segmented object instances in COCO dataset. 
The object masks in COCO dataset are given by polygons. To obtain more accurate object masks, we preprocess the COCO object masks using a segmentation refinement method~\cite{cheng2020cascadepsp}. 
Places2 dataset is a large-scale dataset for scene recognition and contains about 10 million images covering more than 205 scene categories. 
For Places2 dataset, since there is no segmentation ground-truth available, we annotate the object masks using a segmentation method~\cite{li2021fully}.

\subsection{Comparison with State-of-the-art Methods}

\subsubsection{Qualitative evaluation}
\begin{figure}[t]
\begin{center}
    \centering
    \includegraphics[width=\textwidth]{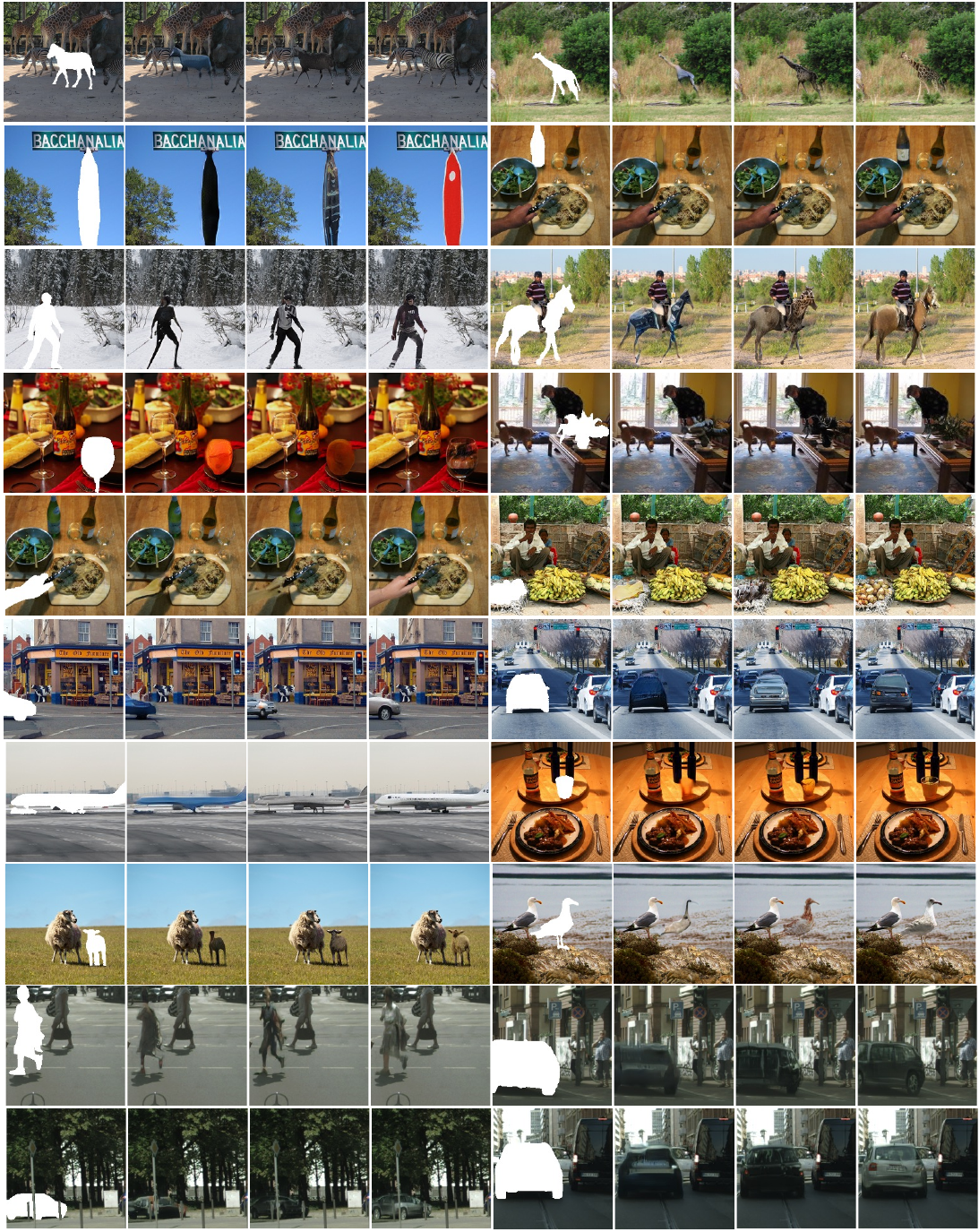}\\
    \scriptsize{\hfill{Input} \hfill\hfill  {DeepFillV2} \hfill\hfill  {CoModGAN} \hfill\hfill {Ours} \hfill\hfill {Input} \hfill\hfill  {DeepFillV2} \hfill\hfill  {CoModGAN} \hfill\hfill {Ours} \hfill}
\caption{Object inpainting results of our method and state-of-the-art methods. Our method can generate objects coherent with the context in terms of both appearance and semantic meanings, while the generated contents of previous approaches seldom resemble reasonable objects. 
}
    \label{fig_results}
    \vspace{-10pt}
\end{center}%
\end{figure}

Fig.~\ref{fig_results} shows the object inpainting results of the proposed method and state-of-the-art methods. Fig.~\ref{fig_diverse} shows the multiple diverse results produced by our method for the same input images. 
Since the existing deep learning based inpainting methods mainly focus on the coherency of appearance between inpainted regions and known regions and only model the bottom-up generation process, they do not perform well for object inpainting. Even when trained on the object datasets, the object inpainting results of the previous approaches are still far from satisfactory. As we can see from the results, DeepFillV2 usually generates a colored shape hardly resembling an object. Benefiting from the powerful StyleGAN architecture, CoModGAN can produce relatively more object-like results, but often without a consistent semantic meaning,~\eg, the horse with giraffe patterns as shown in the right column of the third row. 
In comparison, our method combines the bottom-up and the top-down generation process to achieve both low-level and high-level coherency between the generated content and the surrounding. 
Our method can generate objects that can naturally blend into the context in the sense of both appearance and semantic meanings. The object appearance is consistent with the environment,~\eg lighting, color, and style, and is also well aligned with the corresponding semantic class. 

\subsubsection{Quantitative Evaluation}
Table~\ref{table_lpips} reports quantitative evaluation results on COCO, Places2, and Cityscapes datasets. The evaluation results show that our method outperforms the state-of-the-art methods on all metrics, especially the significantly lower FID. 
Since FID measures the distance between the distribution of the deep features of generated images and real images, the lower FID scores imply that the objects generated by our model have a closer distribution to the distribution of natural objects. 
This further demonstrates the superiority of our method in terms of object inpainting. 
\begin{table}[t]
\caption{\small Quantitative evaluation results. }
\vspace{-0pt}
\label{table_lpips}
\small
\begin{center}
\begin{tabular}{c||cc|cc|cc}
\hline
 & \multicolumn{2}{c|}{COCO} &\multicolumn{2}{c|}{Places2} &\multicolumn{2}{c}{Cityscapes}\\
% \hline
Method&FID &LPIPS &FID &LPIPS &FID &LPIPS\\
\hline
CoModGAN    &7.693&0.1122 &7.471&0.1086 &8.161&0.0491\\
DeepFillV2  &10.56&0.1216 &8.751&0.1201      &10.56&0.0542\\
RFR         &13.38&0.1141 &14.22&0.1125      &15.92&0.0497\\
Ours        &\textbf{4.700}&\textbf{0.1049} &\textbf{3.801}&\textbf{0.0928} &\textbf{7.411}&\textbf{0.0458}\\
\hline
\end{tabular}
\end{center}
\vspace{-0pt}
\end{table}
% \begin{table}[t]
% % \setlength{\tabcolsep}{2pt}
% \caption{\small Places. }
% \vspace{-0pt}
% \label{table_places}
% \small
% \begin{center}
% \begin{tabular}{c||ccccc}
% \hline
% Method& L1 Error &PSNR &SSIM &FID &LPIPS\\
% \hline
% CoModGAN & .0373 & 21.66 & .8587 &7.693&0.1122\\
% DeepFillV2 & .0391 & 21.08 & .8569&10.56&0.1216\\
% % RFR &   &   &  &13.38&\\
% Ours &.0360 & 22.27 & .8690&4.700&0.1049\\
% \hline
% \end{tabular}
% COCO\\
% \begin{tabular}{c||cccc}
% \hline
% Method& L1 Error &PSNR &SSIM &FID\\
% \hline
% CoModGAN &0.0130 & 28.51 & 0.9390&7.471\\  
% DeepFillV2 &0.0135 & 28.22 & 0.9387& \\
% % RFR &0.0105 & 29.48 & 0.9426&  \\
% Ours &0.0125 & 28.70 & 0.9432 &3.801\\
% \hline
% \end{tabular}
% Places\\
% \begin{tabular}{c||cccc}
% \hline
% Method& L1 Error &PSNR &SSIM &FID\\
% \hline
% CoModGAN &0.0130 & 28.51 & 0.9390&8.161\\  
% DeepFillV2 &0.0135 & 28.22 & 0.9387&10.303\\
% % RFR &0.0105 & 29.48 & 0.9426&  15.926\\
% Ours &0.0125 & 28.70 & 0.9432 &7.411\\
% \hline
% \end{tabular}
% Cityscapes\\
% \end{center}
% \vspace{-0pt}
% \end{table}

\subsection{Ablation Study}
\begin{figure}[t]
\begin{center}
    \centering
    \includegraphics[width=.9\textwidth]{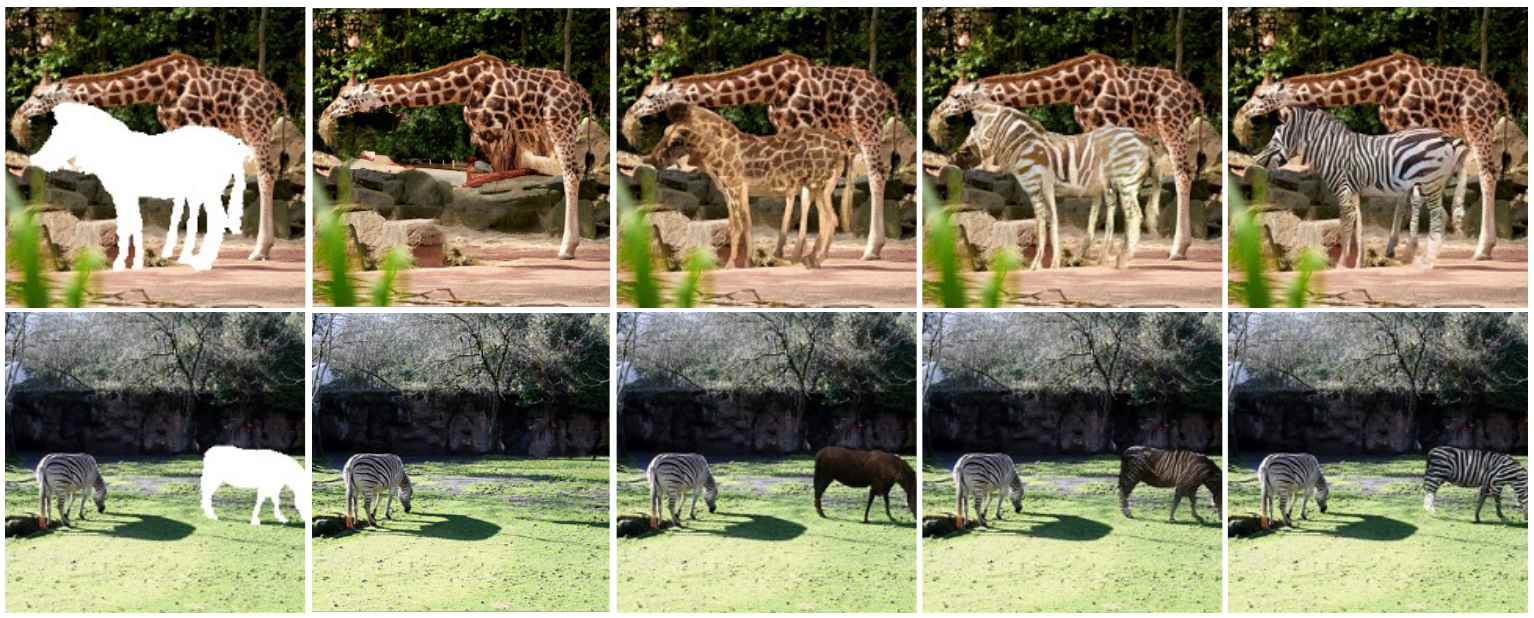}
    % \scriptsize{\hfill{Images} \hfill\hfill  {Sketches} \hfill\hfill  {Results} \hfill\hfill {Images} \hfill\hfill {Sketches} \hfill\hfill {Results} \hfill}
\caption{From left to right are: (a) input, (b) without object training data, (c) without predictive class embedding, (d) without top-down stream, (e) full model. }
    \label{fig_ablation}
    \vspace{-10pt}
\end{center}%
\end{figure}

\begin{figure}[t]
\begin{center}
    \centering
    \includegraphics[width=.9\textwidth]{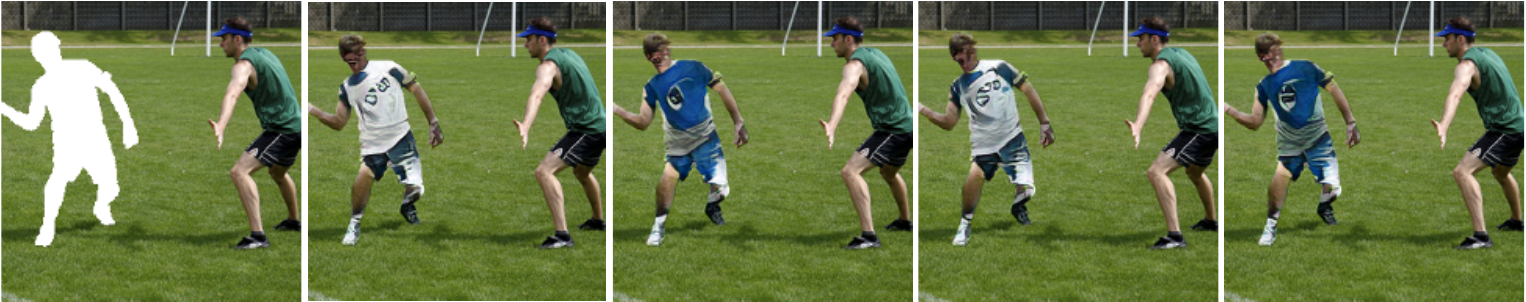}\\
    \includegraphics[width=.9\textwidth]{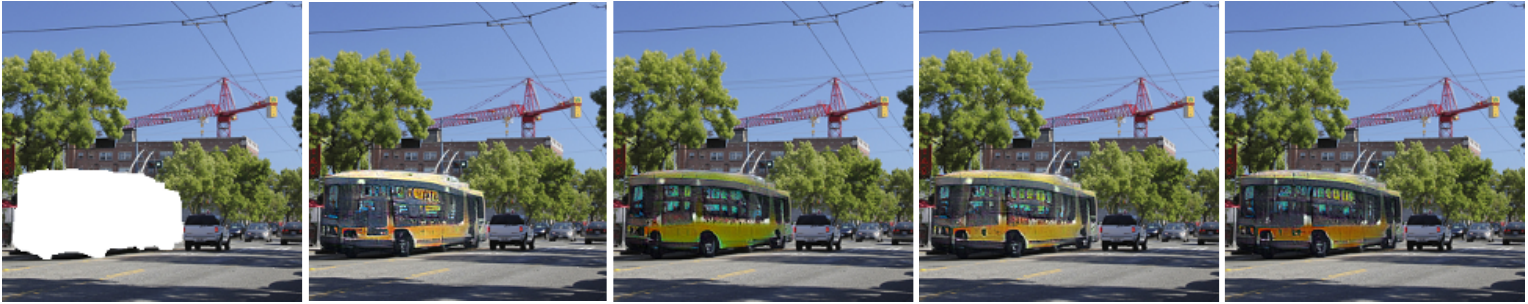}\\
    \includegraphics[width=.9\textwidth]{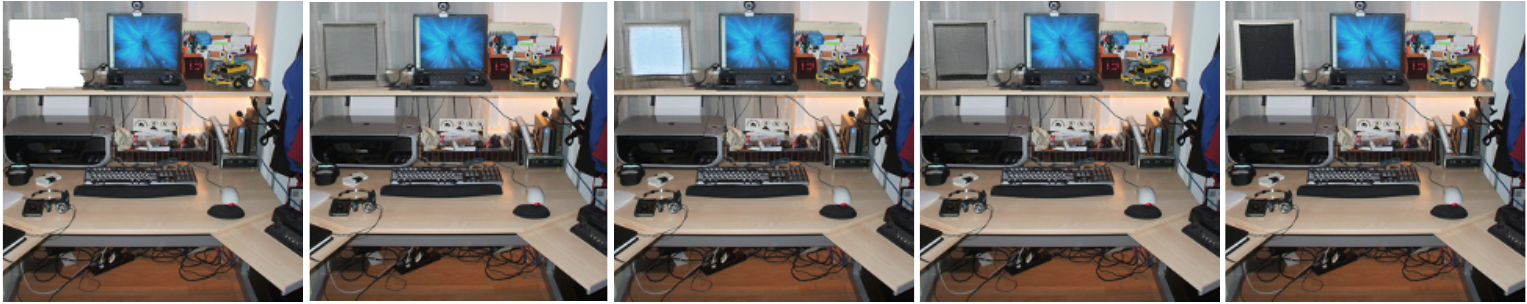}\\
    \includegraphics[width=.9\textwidth]{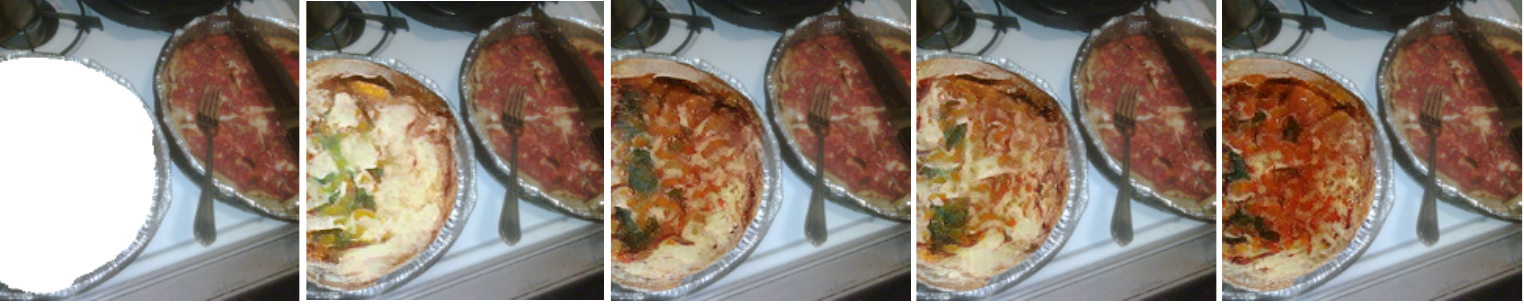}\\
    \includegraphics[width=.9\textwidth]{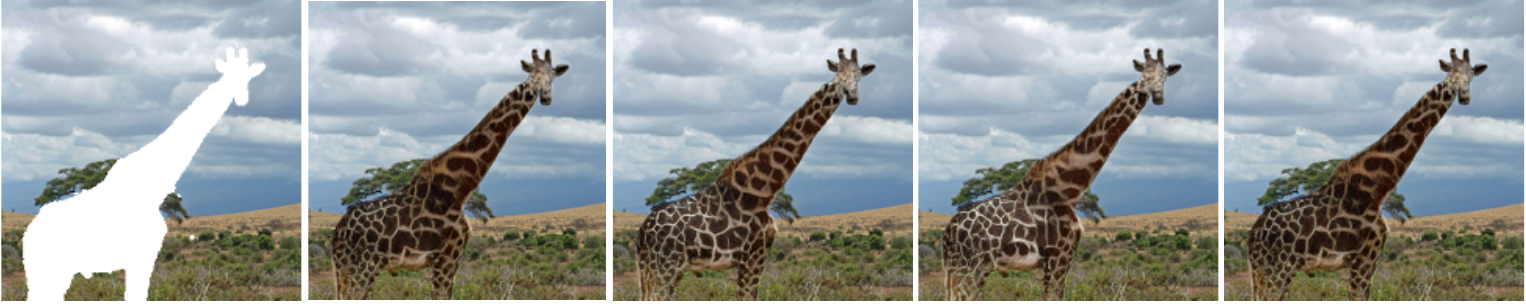}\\
    % \scriptsize{\hfill{Input} \hfill\hfill  {DeepFillV2} \hfill\hfill  {CoModGAN} \hfill\hfill {Ours} \hfill\hfill {Input} \hfill\hfill  {DeepFillV2} \hfill\hfill  {CoModGAN} \hfill\hfill {Ours} \hfill}
\caption{Our method can produce multiple diverse object inpainting results for the same input image by using different random latent code $z$. }
    \label{fig_diverse}
    \vspace{-10pt}
\end{center}%
\end{figure}

In this section, we discuss the effect of each component. First, different from previous work on image inpainting which generates the training data using random masks, we construct the specialized training data for object inpainting to incorporate object prior. Without this prior, the trained inpainting model usually has the bias towards background generation and will not generate objects when filling a missing region, as shown in Fig.~\ref{fig_ablation} (b). The predictive class embedding (PCE) extracts class-related information from the context. Without this module, the model trained on object data might be able to produce object-like content. However, it is challenging to generate a semantically reasonable object without knowing the object's class. As shown in Fig.~\ref{fig_ablation} (c), usually the appearance of the generated objects are simply taken from the nearby regions. For instance, in the second row, the model without PCE generates an object of zebra shape but with the texture of a nearby giraffe. 
The top-down stream takes the semantic object mask as input, which provides stronger spatial semantic guidance for object generation. Without this information, the model can only access class-related information from PCE, which is insufficient for hallucinating object appearance. Hence the model will still rely on the appearance of the surrounding area. As shown in Fig.~\ref{fig_ablation} (d), although the model without the top-down stream can produce some zebra strikes, the color of the zebra seems to be from the surrounding background area. Table~\ref{table_ablation_score} reports FID scores with and without each component. We can see that the predictive class embedding and the incorporation of the top-down stream can significantly reduce the FID by providing class-related information. %The third and the fourth row compare the results obtained with and without SC AdaIN. To show the effect of SC AdaIN, we 
\begin{table}[t]
\caption{\small Effect of each component in terms of FID and LPIPS. }
% \vspace{-0pt}
\label{table_ablation_score}
\small
\begin{center}
\begin{tabular}{cccc||cc}
\hline
& Object Data & PCE & Top-down &FID&LPIPS\\
\hline
% $\surd$ &         &         & &               &  \\
        & $\surd$ &         &              &6.144 &0.1066\\
        & $\surd$ & $\surd$ &           &5.434 &0.1081\\
        % & $\surd$ & $\surd$ &  $\surd$ &       &4.732\\%899
        & $\surd$ & $\surd$ & $\surd$ &4.700 &0.1049\\%769
\hline
\end{tabular}
\end{center}
\vspace{-0pt}
\end{table}

\section{Conclusion and Future Work}
We study a new image inpainting task,~\ie shape-guided object inpainting. We find that existing image inpainting methods are not suitable for object inpainting due to the bias towards background and a lack of top-down guidance. Therefore, we design a new data preparation method that incorporates object priors by using object instances as holes and propose a Contextual Object Generator (CogNet) with a two-stream network architecture that combines the bottom-up image completion process with a top-down object generation process. 
Experiments demonstrate that the proposed method can generate realistic objects that fit the context in terms of both visual appearance and semantic meanings. 

% In this work, the guidance information for inpainting is implicitly given by the hole shape. A possible future research direction is to incorporate explicit guidance into this framework. In addition, 
The proposed method can be easily extended to inpaint partially missing objects by using partial instances masks in training. This can be an interesting topic for future work. 
\clearpage
% ---- Bibliography ----
%
% BibTeX users should specify bibliography style 'splncs04'.
% References will then be sorted and formatted in the correct style.
%
\bibliographystyle{splncs04}
\bibliography{egbib}

\begin{thebibliography}{10}
\providecommand{\url}[1]{\texttt{#1}}
\providecommand{\urlprefix}{URL }
\providecommand{\doi}[1]{https://doi.org/#1}

\bibitem{albahar2019guided}
AlBahar, B., Huang, J.B.: Guided image-to-image translation with bi-directional
  feature transformation. In: International Conference on Computer Vision. pp.
  9016--9025 (2019)

\bibitem{ardino2021semantic}
Ardino, P., Liu, Y., Ricci, E., Lepri, B., De~Nadai, M.: Semantic-guided
  inpainting network for complex urban scenes manipulation. IEEE (2021)

\bibitem{ballester2001filling}
Ballester, C., Bertalmio, M., Caselles, V., Sapiro, G., Verdera, J.: Filling-in
  by joint interpolation of vector fields and gray levels. IEEE Transaction on
  Image Process.  \textbf{10}(8),  1200--1211 (2001)

\bibitem{barnes2009patchmatch}
Barnes, C., Shechtman, E., Finkelstein, A., Goldman, D.B.: Patchmatch: A
  randomized correspondence algorithm for structural image editing. ACM
  Transactions on Graphics  \textbf{28}(3), ~24 (2009)

\bibitem{10.1145/344779.344972}
Bertalmio, M., Sapiro, G., Caselles, V., Ballester, C.: Image inpainting. In:
  Proceedings of the 27th Annual Conference on Computer Graphics and
  Interactive Techniques. p. 417–424. SIGGRAPH '00, ACM Press/Addison-Wesley
  Publishing Co., USA (2000). \doi{10.1145/344779.344972},
  \url{https://doi.org/10.1145/344779.344972}

\bibitem{brock2018large}
Brock, A., Donahue, J., Simonyan, K.: Large scale gan training for high
  fidelity natural image synthesis. arXiv preprint arXiv:1809.11096  (2018)

\bibitem{chen2017photographic}
Chen, Q., Koltun, V.: Photographic image synthesis with cascaded refinement
  networks. pp. 1511--1520 (2017)

\bibitem{cheng2020cascadepsp}
Cheng, H.K., Chung, J., Tai, Y.W., Tang, C.K.: Cascadepsp: Toward
  class-agnostic and very high-resolution segmentation via global and local
  refinement. In: Proceedings of the IEEE/CVF Conference on Computer Vision and
  Pattern Recognition. pp. 8890--8899 (2020)

\bibitem{Cordts2016Cityscapes}
Cordts, M., Omran, M., Ramos, S., Rehfeld, T., Enzweiler, M., Benenson, R.,
  Franke, U., Roth, S., Schiele, B.: The cityscapes dataset for semantic urban
  scene understanding. In: Proc. of the IEEE Conference on Computer Vision and
  Pattern Recognition (CVPR) (2016)

\bibitem{drori2003fragment}
Drori, I., Cohen-Or, D., Yeshurun, H.: Fragment-based image completion. In: ACM
  Transactions on Graphics, pp. 303--312 (2003)

\bibitem{efros1999texture}
Efros, A.A., Leung, T.K.: Texture synthesis by non-parametric sampling. In:
  International Conference on Computer Vision. vol.~2, pp. 1033--1038. IEEE
  (1999)

\bibitem{goodfellow2014generative}
Goodfellow, I., Pouget-Abadie, J., Mirza, M., Xu, B., Warde-Farley, D., Ozair,
  S., Courville, A., Bengio, Y.: Generative adversarial nets. Conference on
  Neural Information Processing Systems  \textbf{27} (2014)

\bibitem{gulrajani2017improved}
Gulrajani, I., Ahmed, F., Arjovsky, M., Dumoulin, V., Courville, A.C.: Improved
  training of wasserstein gans. Advances in neural information processing
  systems  \textbf{30} (2017)

\bibitem{heusel2017gans}
Heusel, M., Ramsauer, H., Unterthiner, T., Nessler, B., Hochreiter, S.: Gans
  trained by a two time-scale update rule converge to a local nash equilibrium.
  Conference on Neural Information Processing Systems  \textbf{30} (2017)

\bibitem{huang2018multimodal}
Huang, X., Liu, M.Y., Belongie, S., Kautz, J.: Multimodal unsupervised
  image-to-image translation. In: European Conference on Computer Vision. pp.
  172--189 (2018)

\bibitem{iizuka2017globally}
Iizuka, S., Simo-Serra, E., Ishikawa, H.: Globally and locally consistent image
  completion. ACM Transactions on Graphics  \textbf{36}(4),  1--14 (2017)

\bibitem{isola2017image}
Isola, P., Zhu, J.Y., Zhou, T., Efros, A.A.: Image-to-image translation with
  conditional adversarial networks. In: IEEE Conference on Computer Vision and
  Pattern Recognition. pp. 1125--1134 (2017)

\bibitem{jo2019sc}
Jo, Y., Park, J.: Sc-fegan: Face editing generative adversarial network with
  user's sketch and color. In: International Conference on Computer Vision. pp.
  1745--1753 (2019)

\bibitem{johnson2016perceptual}
Johnson, J., Alahi, A., Fei-Fei, L.: Perceptual losses for real-time style
  transfer and super-resolution. In: European Conference on Computer Vision.
  pp. 694--711. Springer (2016)

\bibitem{joung2012reliable}
Joung, J.H., Ryoo, M.S., Choi, S., Kim, S.R.: Reliable object detection and
  segmentation using inpainting. In: 2012 IEEE/RSJ International Conference on
  Intelligent Robots and Systems. pp. 3871--3876. IEEE (2012)

\bibitem{karras2017progressive}
Karras, T., Aila, T., Laine, S., Lehtinen, J.: Progressive growing of gans for
  improved quality, stability, and variation. arXiv preprint arXiv:1710.10196
  (2017)

\bibitem{karras2019style}
Karras, T., Laine, S., Aila, T.: A style-based generator architecture for
  generative adversarial networks. In: IEEE Conference on Computer Vision and
  Pattern Recognition (2019)

\bibitem{karras2020analyzing}
Karras, T., Laine, S., Aittala, M., Hellsten, J., Lehtinen, J., Aila, T.:
  Analyzing and improving the image quality of stylegan. In: IEEE Conference on
  Computer Vision and Pattern Recognition. pp. 8110--8119 (2020)

\bibitem{katircioglu2019self}
Katircioglu, I., Rhodin, H., Constantin, V., Sp{\"o}rri, J., Salzmann, M., Fua,
  P.: Self-supervised training of proposal-based segmentation via background
  prediction. arXiv preprint arXiv:1907.08051  (2019)

\bibitem{katircioglu2020self}
Katircioglu, I., Rhodin, H., Constantin, V., Sp{\"o}rri, J., Salzmann, M., Fua,
  P.: Self-supervised segmentation via background inpainting. arXiv preprint
  arXiv:2011.05626  (2020)

\bibitem{ke2021occlusion}
Ke, L., Tai, Y.W., Tang, C.K.: Occlusion-aware video object inpainting. In:
  International Conference on Computer Vision. pp. 14468--14478 (2021)

\bibitem{krizhevsky2012imagenet}
Krizhevsky, A., Sutskever, I., Hinton, G.E.: Imagenet classification with deep
  convolutional neural networks. Advances in neural information processing
  systems  \textbf{25} (2012)

\bibitem{kwatra2005texture}
Kwatra, V., Essa, I., Bobick, A., Kwatra, N.: Texture optimization for
  example-based synthesis. In: ACM Trans. Graphic., pp. 795--802 (2005)

\bibitem{lahiri2020prior}
Lahiri, A., Jain, A.K., Agrawal, S., Mitra, P., Biswas, P.K.: Prior guided gan
  based semantic inpainting. In: IEEE Conference on Computer Vision and Pattern
  Recognition (2020)

\bibitem{li2020recurrent}
Li, J., Wang, N., Zhang, L., Du, B., Tao, D.: Recurrent feature reasoning for
  image inpainting. In: IEEE Conference on Computer Vision and Pattern
  Recognition. pp. 7760--7768 (2020)

\bibitem{li2021fully}
Li, Y., Zhao, H., Qi, X., Wang, L., Li, Z., Sun, J., Jia, J.: Fully
  convolutional networks for panoptic segmentation. In: Proceedings of the
  IEEE/CVF Conference on Computer Vision and Pattern Recognition. pp. 214--223
  (2021)

\bibitem{liao2021image}
Liao, L., Xiao, J., Wang, Z., Lin, C.W., Satoh, S.: Image inpainting guided by
  coherence priors of semantics and textures. In: IEEE Conference on Computer
  Vision and Pattern Recognition (2021)

\bibitem{lin2014microsoft}
Lin, T.Y., Maire, M., Belongie, S., Hays, J., Perona, P., Ramanan, D.,
  Doll{\'a}r, P., Zitnick, C.L.: Microsoft coco: Common objects in context. In:
  European conference on computer vision. pp. 740--755. Springer (2014)

\bibitem{liu2018image}
Liu, G., Reda, F.A., Shih, K.J., Wang, T.C., Tao, A., Catanzaro, B.: Image
  inpainting for irregular holes using partial convolutions. In: European
  Conference on Computer Vision (2018)

\bibitem{liu2020rethinking}
Liu, H., Jiang, B., Song, Y., Huang, W., Yang, C.: Rethinking image inpainting
  via a mutual encoder-decoder with feature equalizations. In: European
  Conference on Computer Vision (2020)

\bibitem{liu2019coherent}
Liu, H., Jiang, B., Xiao, Y., Yang, C.: Coherent semantic attention for image
  inpainting. In: IEEE International Conference on Computer Vision (2019)

\bibitem{liu2021deflocnet}
Liu, H., Wan, Z., Huang, W., Song, Y., Han, X., Liao, J., Jiang, B., Liu, W.:
  Deflocnet: Deep image editing via flexible low-level controls. In: IEEE
  Conference on Computer Vision and Pattern Recognition (2021)

\bibitem{lucic2018gans}
Lucic, M., Kurach, K., Michalski, M., Gelly, S., Bousquet, O.: Are gans created
  equal? a large-scale study. In: Conference on Neural Information Processing
  Systems (2018)

\bibitem{nazeri2019edgeconnect}
Nazeri, K., Ng, E., Joseph, T., Qureshi, F.Z., Ebrahimi, M.: Edgeconnect:
  Generative image inpainting with adversarial edge learning. arXiv preprint
  arXiv:1901.00212  (2019)

\bibitem{park2019semantic}
Park, T., Liu, M.Y., Wang, T.C., Zhu, J.Y.: Semantic image synthesis with
  spatially-adaptive normalization. In: IEEE Conference on Computer Vision and
  Pattern Recognition. pp. 2337--2346 (2019)

\bibitem{NEURIPS2019_9015}
Paszke, A., Gross, S., Massa, F., Lerer, A., Bradbury, J., Chanan, G., Killeen,
  T., Lin, Z., Gimelshein, N., Antiga, L., Desmaison, A., Kopf, A., Yang, E.,
  DeVito, Z., Raison, M., Tejani, A., Chilamkurthy, S., Steiner, B., Fang, L.,
  Bai, J., Chintala, S.: Pytorch: An imperative style, high-performance deep
  learning library. In: Wallach, H., Larochelle, H., Beygelzimer, A.,
  d\textquotesingle Alch\'{e}-Buc, F., Fox, E., Garnett, R. (eds.) Advances in
  Neural Information Processing Systems 32, pp. 8024--8035. Curran Associates,
  Inc. (2019),
  \url{http://papers.neurips.cc/paper/9015-pytorch-an-imperative-style-high-performance-deep-learning-library.pdf}

\bibitem{pathak2016context}
Pathak, D., Krahenbuhl, P., Donahue, J., Darrell, T., Efros, A.A.: Context
  encoders: Feature learning by inpainting. In: IEEE Conference on Computer
  Vision and Pattern Recognition (2016)

\bibitem{portenier2018faceshop}
Portenier, T., Hu, Q., Szabo, A., Bigdeli, S.A., Favaro, P., Zwicker, M.:
  Faceshop: Deep sketch-based face image editing. ACM Transactions on Graphics
  (2018)

\bibitem{radford2015unsupervised}
Radford, A., Metz, L., Chintala, S.: Unsupervised representation learning with
  deep convolutional generative adversarial networks. arXiv preprint
  arXiv:1511.06434  (2015)

\bibitem{ren2019structureflow}
Ren, Y., Yu, X., Zhang, R., Li, T.H., Liu, S., Li, G.: Structureflow: Image
  inpainting via structure-aware appearance flow. In: International Conference
  on Computer Vision (2019)

\bibitem{shen2019towards}
Shen, Z., Huang, M., Shi, J., Xue, X., Huang, T.S.: Towards instance-level
  image-to-image translation. In: IEEE Conference on Computer Vision and
  Pattern Recognition. pp. 3683--3692 (2019)

\bibitem{song2018contextual}
Song, Y., Yang, C., Lin, Z., Liu, X., Huang, Q., Li, H., Kuo, C.C.J.:
  Contextual-based image inpainting: Infer, match, and translate. In: European
  Conference on Computer Vision. pp. 3--19 (2018)

\bibitem{suin2021distillation}
Suin, M., Purohit, K., Rajagopalan, A.: Distillation-guided image inpainting.
  In: International Conference on Computer Vision. pp. 2481--2490 (2021)

\bibitem{wang2018high}
Wang, T.C., Liu, M.Y., Zhu, J.Y., Tao, A., Kautz, J., Catanzaro, B.:
  High-resolution image synthesis and semantic manipulation with conditional
  gans. pp. 8798--8807 (2018)

\bibitem{wangimage}
Wang, Y., Tao, X., Qi, X., Shen, X., Jia, J.: Image inpainting via generative
  multi-column convolutional neural networks. In: Conference on Neural
  Information Processing Systems (2018)

\bibitem{xiao2019cisi}
Xiao, J., Liao, L., Liu, Q., Hu, R.: Cisi-net: Explicit latent content
  inference and imitated style rendering for image inpainting. In: he AAAI
  Conference on Artificial Intelligence (2019)

\bibitem{xiao2020noise}
Xiao, K.Y., Engstrom, L., Ilyas, A., Madry, A.: Noise or signal: The role of
  image backgrounds in object recognition. In: International Conference on
  Learning Representations (2020)

\bibitem{xiong2019foreground}
Xiong, W., Yu, J., Lin, Z., Yang, J., Lu, X., Barnes, C., Luo, J.:
  Foreground-aware image inpainting. In: IEEE Conference on Computer Vision and
  Pattern Recognition (2019)

\bibitem{yang2017high}
Yang, C., Lu, X., Lin, Z., Shechtman, E., Wang, O., Li, H.: High-resolution
  image inpainting using multi-scale neural patch synthesis. In: IEEE
  Conference on Computer Vision and Pattern Recognition (2017)

\bibitem{yang2020learning}
Yang, J., Qi, Z., Shi, Y.: Learning to incorporate structure knowledge for
  image inpainting. In: he AAAI Conference on Artificial Intelligence (2020)

\bibitem{yang2020deep}
Yang, S., Wang, Z., Liu, J., Guo, Z.: Deep plastic surgery: Robust and
  controllable image editing with human-drawn sketches. In: European Conference
  on Computer Vision (2020)

\bibitem{yi2020contextual}
Yi, Z., Tang, Q., Azizi, S., Jang, D., Xu, Z.: Contextual residual aggregation
  for ultra high-resolution image inpainting. In: IEEE Conference on Computer
  Vision and Pattern Recognition (2020)

\bibitem{yu2018generative}
Yu, J., Lin, Z., Yang, J., Shen, X., Lu, X., Huang, T.S.: Generative image
  inpainting with contextual attention. In: IEEE Conference on Computer Vision
  and Pattern Recognition (2018)

\bibitem{yu2019free}
Yu, J., Lin, Z., Yang, J., Shen, X., Lu, X., Huang, T.S.: Free-form image
  inpainting with gated convolution. In: International Conference on Computer
  Vision (2019)

\bibitem{yu2020region}
Yu, T., Guo, Z., Jin, X., Wu, S., Chen, Z., Li, W., Zhang, Z., Liu, S.: Region
  normalization for image inpainting. In: he AAAI Conference on Artificial
  Intelligence (2020)

\bibitem{zeng2019learning}
Zeng, Y., Fu, J., Chao, H., Guo, B.: Learning pyramid-context encoder network
  for high-quality image inpainting. In: IEEE Conference on Computer Vision and
  Pattern Recognition. pp. 1486--1494 (2019)

\bibitem{zeng2020high}
Zeng, Y., Lin, Z., Yang, J., Zhang, J., Shechtman, E., Lu, H.: High-resolution
  image inpainting with iterative confidence feedback and guided upsampling.
  In: European Conference on Computer Vision. Springer (2020)

\bibitem{zhang2020text}
Zhang, L., Chen, Q., Hu, B., Jiang, S.: Text-guided neural image inpainting.
  In: Proceedings of the 28th ACM International Conference on Multimedia. pp.
  1302--1310 (2020)

\bibitem{zhang2018perceptual}
Zhang, R., Isola, P., Efros, A.A., Shechtman, E., Wang, O.: The unreasonable
  effectiveness of deep features as a perceptual metric. In: IEEE Conference on
  Computer Vision and Pattern Recognition (2018)

\bibitem{zhang2019residual}
Zhang, Y., Li, K., Li, K., Zhong, B., Fu, Y.: Residual non-local attention
  networks for image restoration. arXiv preprint arXiv:1903.10082  (2019)

\bibitem{zhao2021comodgan}
Zhao, S., Cui, J., Sheng, Y., Dong, Y., Liang, X., Chang, E.I., Xu, Y.: Large
  scale image completion via co-modulated generative adversarial networks. In:
  International Conference on Learning Representations (ICLR) (2021)

\bibitem{zhao2019guided}
Zhao, Y., Price, B., Cohen, S., Gurari, D.: Guided image inpainting: Replacing
  an image region by pulling content from another image. In: 2019 IEEE Winter
  Conference on Applications of Computer Vision (WACV). pp. 1514--1523. IEEE
  (2019)

\bibitem{zhou2017places}
Zhou, B., Lapedriza, A., Khosla, A., Oliva, A., Torralba, A.: Places: A 10
  million image database for scene recognition. IEEE Transactions on Pattern
  Analysis and Machine Intelligence  \textbf{40}(6),  1452--1464 (2017)

\bibitem{zhou2021transfill}
Zhou, Y., Barnes, C., Shechtman, E., Amirghodsi, S.: Transfill:
  Reference-guided image inpainting by merging multiple color and spatial
  transformations. In: IEEE Conference on Computer Vision and Pattern
  Recognition. pp. 2266--2276 (2021)

\bibitem{zhu2017unpaired}
Zhu, J.Y., Park, T., Isola, P., Efros, A.A.: Unpaired image-to-image
  translation using cycle-consistent adversarial networks. In: International
  Conference on Computer Vision. pp. 2223--2232 (2017)

\bibitem{zhu2016object}
Zhu, Z., Xie, L., Yuille, A.L.: Object recognition with and without objects.
  In: International Joint Conference on Artificial Intelligence (2017)

\end{thebibliography}
\end{document}